\documentclass[10pt,twocolumn,letterpaper]{article}

\usepackage{cvpr}              

\usepackage{graphicx}
\usepackage{amsmath}
\usepackage{amssymb}
\usepackage{booktabs}
\usepackage{multirow}
\usepackage{arydshln} 
\usepackage{multirow}
\usepackage[accsupp]{axessibility}

\usepackage[pagebackref,breaklinks,colorlinks]{hyperref}

\usepackage{times}
\usepackage{epsfig}
\usepackage{amsmath}
\usepackage{amssymb}
\usepackage{mathptmx}

\usepackage[capitalize]{cleveref}
\crefname{section}{Sec.}{Secs.}
\Crefname{section}{Section}{Sections}
\Crefname{table}{Table}{Tables}
\crefname{table}{Tab.}{Tabs.}

\def\cvprPaperID{3170} 
\def\confName{CVPR}
\def\confYear{2022}

\begin{document}

\title{XMP-Font: Self-Supervised Cross-Modality Pre-training for \\ Few-Shot Font Generation}

\author{Wei Liu\footnotemark[1]~~~~~~~~~~~Fangyue Liu\footnotemark[1]~~~~~~~~~~~Fei Ding~~~~~~~~~~~Qian He~~~~~~~~~~~Zili Yi\footnotemark[2]\\
ByteDance Ltd, Beijing, China\\
{\tt\small liujikun63@gmail.com}~~~~~~~~~~~~~{\tt\small liufangyue1999@hotmail.com}~~~~~~~~~~~~~~~~{\tt\small dingfei.212@bytedance.com}\\
~~~~~~~{\tt\small heqian@bytedance.com}~~~~~~~~~~~~{\tt\small yizili14@gmail.com}\\
}

\twocolumn[{%
\renewcommand\twocolumn[1][]{#1}%
\maketitle
\begin{center}
 \centering
\captionsetup{type=figure}
      \includegraphics[width=\textwidth]{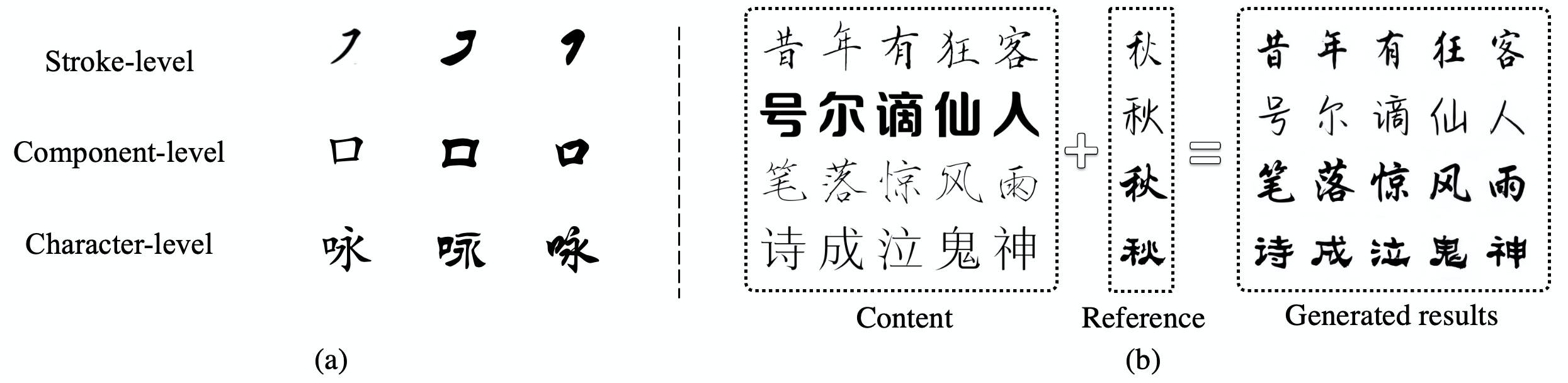}
        \caption[figure]{(a) Illustrations of different Chinese font styles. Typically, a font style involves unique morphological structures of a character at multiple scales. For example, stroke-level styles involve stroke-level features such as weight, hollowness and serif-ness. Component-level styles implies how strokes are oriented and joined to form a component. Character-level styles engage character-level features including the component layout, inter-component spacing and ``connected-stroke’’. (b) Exemplar few-shot font generation results of our method. Please note that each row presents a disparate font style by using only one glyph as reference. In addition, our model generates faithful and consistent results regardless of the type of the source font or the reference variant. \label{fig:teaser}}
\end{center}
}]

\renewcommand{\thefootnote}{\fnsymbol{footnote}}
\footnotetext[1]{These authors contributed equally and should be considered co-first authors.}

\footnotetext[2]{Corresponding author.}
\begin{abstract}
Generating a new font library is a very labor-intensive and time-consuming job for glyph-rich scripts. Few-shot font generation is thus required, as it requires only a few glyph references without fine-tuning during test.
Existing methods follow the style-content disentanglement paradigm and expect novel fonts to be produced by combining the style codes of the reference glyphs and the content representations of the source.
However, these few-shot font generation methods either fail to capture content-independent style representations, or employ localized component-wise style representations, which is insufficient to model many Chinese font styles that involve hyper-component features such as inter-component spacing and ``connected-stroke". To resolve these drawbacks and make the style representations more reliable, we propose a self-supervised cross-modality pre-training strategy and a cross-modality transformer-based encoder that is conditioned jointly on the glyph image and the corresponding stroke labels. The cross-modality encoder is pre-trained in a self-supervised manner to allow effective capture of cross- and intra-modality correlations, which facilitates the content-style disentanglement and modeling style representations of all scales (stroke-level, component-level and character-level). The pre-trained encoder is then applied to the downstream font generation task without fine-tuning. Experimental comparisons of our method with state-of-the-art methods demonstrate our method successfully transfers styles of all scales. In addition, it only requires one reference glyph and achieves the lowest rate of bad cases in the few-shot font generation task (28\% lower than the second best).
\end{abstract}

\section{Introduction}

The few-shot font generation task (FFG) aims to produce a new font library using only a few glyphs as reference, without additional fine-tuning of the model at the testing stage. FFG is especially a desirable task when designing a new font library for glyph-rich scripts such as Chinese (the total number of characters exceeds 80,000), as the traditional manual font design process is very laborious. FFG is also desired when the target style glyphs are too rare to collect (e.g., historical handwriting). 

Since font styles are highly complex and fine-grained, a simple analysis of the low-level textures of a few reference examples is impossible to perform successful style transfer as in \cite{gatys2016image,huang2017arbitrary,li2017universal,li2018closed,luan2017deep}. A common paradigm used for FFG is to disentangle font-specific style and content information from the given glyphs, and synthesize a new glyph by combining the style embeddings extracted from the reference set and the content representations of the source glyph \cite{cha2020few,gao2019artistic,park2020few,zhang2018separating,sun2017learning,azadi2018multi,park2021multiple,li2021few,srivatsan2019deep,xie2021dg}. Early attempts of this stream~\cite{zhang2018separating,gao2019artistic} employ the universal style representations, using a simple convolutional encoder to extract style embeddings directly from the reference glyph images. However, the universal style representations show limited capabilities in capturing reliable and content-independent style representations due to limited awareness of the character structures and correlations between different regions of the input glyph. More advanced architectures such as DM-Font \cite{cha2020few}, LF-Font \cite{park2020few}, MX-Font \cite{park2021multiple} propose to use structure-aware style representations and learn the localized component-wise style representations.

To make the localized style representation possible, these methods either condition the style encoders jointly upon the glyph image and the corresponding component labels or introduce component-label-guided losses to train the style encoder.  The structure-aware localized style representations remarkably improve the reliability of the style representations. However, as mentioned in \cite{park2020few}, learning component-wise styles solely is insufficient for component-rich glyphs like Chinese characters that have over 200 different types of components. It is hard to cover all component types with a few reference glyphs during test. To relieve this problem, LF-Font \cite{park2020few} simplifies the component-wise styles by a product of component factor and style factor, inspired by low-rank matrix factorization. MX-Font \cite{park2021multiple} extracts multiple style features not explicitly conditioned on component labels, but automatically by multiple experts to represent different local concepts, thus enabling the model to be generalized to a character with unseen components.

Such solutions relieve the ``unseen components’’ issue to some extent. However, they are prone to generating bad cases when failing to generalize the unseen component styles from seen components. On the other hand, component-wise style representations are incapable of capturing character-level style features (e.g., inter-component spacing), which is an important perspective in many Chinese font libraries: see Figure \ref{fig:teaser} (a) for the Chinese font styles of all three scales. 

To address these issues, we make two significant changes. First, we employ the stroke labels rather than the component labels as the atomic representation of character structure, as the stroke set used in Chinese is significantly fewer (about 28) than that of the component set (more than 200), which can be easier to cover with a few reference glyphs or generalized from seen strokes. On the other hand, to enhance the awareness of the stroke-level styles while not losing component-level or character-level style features, we propose to use the unified all-scale style representations instead of the localized component- or stroke-wise styles. This can be achieved by introducing a cross-modality transformer-based encoder that is conditioned jointly on the glyph image and the corresponding stroke labels. On one hand, the self-attention layers used in the encoder is good at capturing both local and global style features. On the other hand, the self-supervised pre-training of the cross-modality encoder inspires the learning the glyph-stroke alignments, which further facilitates the content-style disentanglement and modeling of style representations at multiple scales in the downstream training phase. 

In addition to the cross-modality pre-training mechanism, we propose a LSTM-based stroke loss and a style-content decoupling network which considers spatial information conservation, to enhance the reliability of the model further. Comprehensive analyses of the experimental results demonstrate our method achieves significantly lower rate of bad cases than prior FFG methods and it can successfully generate novel glyphs based on only one reference example.

To sum up, the major contributions of the paper include:
\begin{itemize}
\item For the first time, we introduce the cross-modality transformer-based encoder and the mechanism of cross-modality pre-training to the FFG task. The self- and cross-attention layers in the transformer-based encoder pre-trained with self-supervised signals help capture local and global style features (stroke-level, component-level and character-level features) and learn the glyph-stroke alignments, thus enhancing the structure-awareness of style representations and facilitating the style-content disentanglement in the downstream FFG task. 
\item We elaborate a style-content decoupling network composed of  Efficient Channel Attention (ECA) modules \cite{2020ECA}, and employ an $8\times8$ feature map instead of a simple average-pooled vector to represent styles or contents with the expect to conserve spatial information, which prove to be effective in increasing the reliability of the model.
\item We also propose a novel stroke loss based on a pre-trained LSTM-based stroke order predictor, to enforce the correct stroke order of the generated glyph instead of the existence of stroke labels only, which proves to benefit the structure preservation and faithful generation of stroke-order-related style features (e.g., ``connected-stroke’’).
\item Experimental results see powerful generalizability of our model to unseen font domains. Our model can perform successful font style transfer with only one reference glyph.
\end{itemize}

\section{Related work}

\subsection{Image-to-image translation}
Image-to-image translation methods \cite{choi2018stargan,choi2020stargan,isola2017image,liu2018unified,zhu2017unpaired,yi2017dualgan,yi2017dualgan,liu2019few} that learn the mapping between domains can be used for cross-domain font generation. For example, StarGAN-v2 \cite{choi2020stargan} proposes to do image-to-image translation across multiple-domain in a unified framework. FUNIT \cite{liu2019few} aims to translate an image to the given reference style while preserving the content without fine-tuning the model during test, which can be used for the FFG task.  In this paper, we have our method compared with StarGAN-v2 as for generating glyphs of seen font domains, and also compared with FUNIT on both seen and unseen font domains.

\subsection{Many-shot font generation methods}
Early font generation methods \cite{tian2017zi2zi,wu2020calligan,jiang2019scfont,huang2020rd,gao2020gan} train the cross-domain translators between different font styles. Some font generation methods \cite{wu2020calligan,jiang2019scfont,huang2020rd,gao2020gan} train a translation model first, and fine-tune the translation model with many reference glyphs of the target style. For example, hundreds of reference glyphs in the target domain are used in \cite{jiang2019scfont}. Despite their remarkable performances, their scenario is very limited because collecting hundreds of glyphs with a coherent style can be very expensive. In this paper, we aim to generate an unseen font library without any expensive fine-tuning or collecting a large number of reference glyphs for that style.

\begin{figure*}
        \centering
        \includegraphics[width=1.01\textwidth]{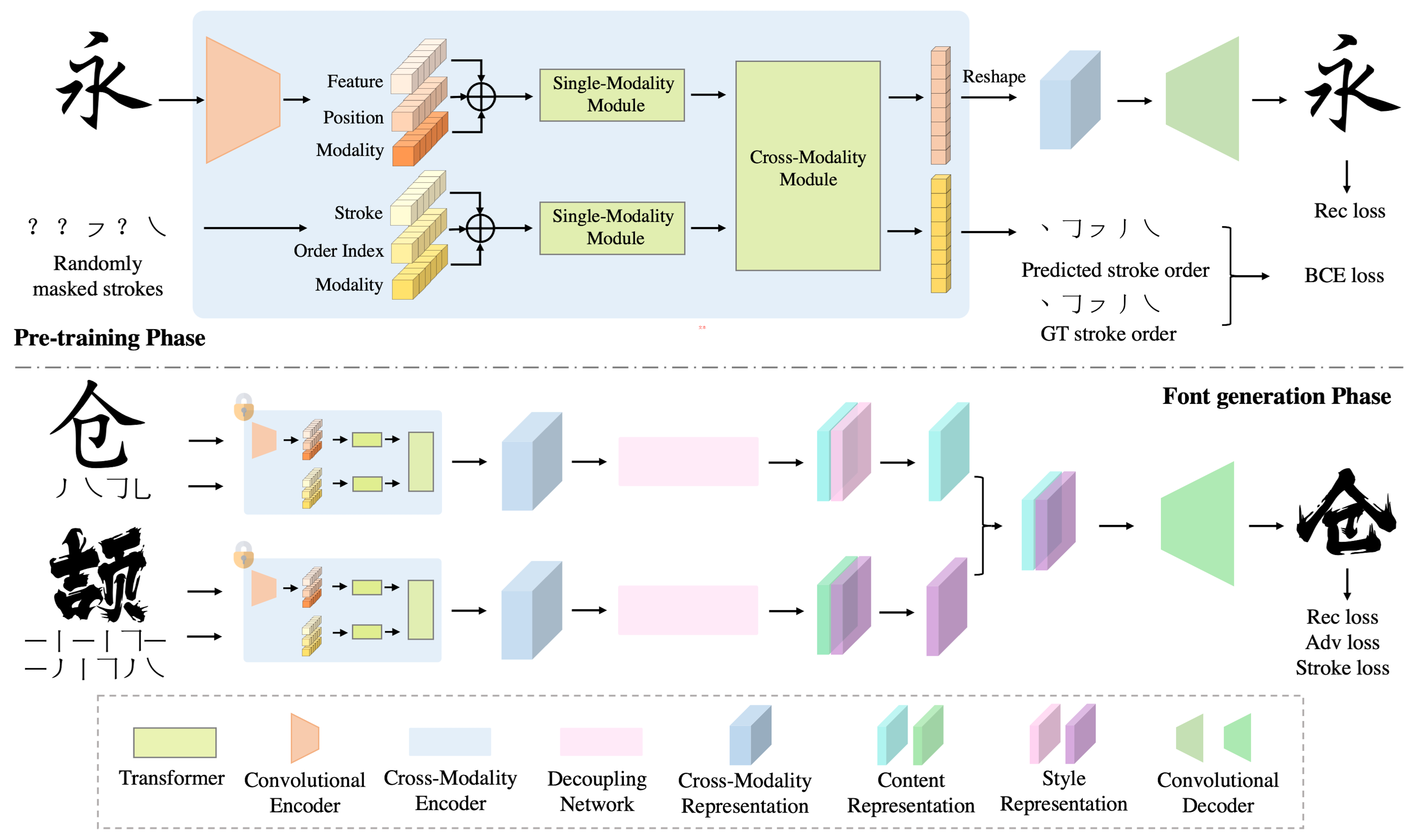}
        \caption[]{The framework of the proposed XMP-Font model. Our approach consists of the pre-trining phase and the downstream font generation phase. The cross-modality encoder is shared in both phases.}
        \label{fig:framework}
\end{figure*}

\subsection{Cross-modality pre-training}

Cross-modality pre-training \cite{tan2019lxmert,DBLPabs210300020,chen2020uniter,li2020oscar,zhang2021vinvl,yu2020ernie,li2021structurallm} is widely used in visual-linguistic tasks such as image-text matching \cite{tan2019lxmert,DBLPabs210300020,chen2020uniter}, visual question answering \cite{su2019vl,tan2019lxmert,li2020oscar,chen2020uniter}, image captioning \cite{zhang2021vinvl}, etc. Cross-modality tasks require the understanding of both modalities, and the alignment and relationships between the two modalities. The pre-training enables the encoder to produce representations with fused cross-modality information, thus benefiting downstream tasks. Motivated by the concept, we introduce such mechanisms to the font generation tasks. The framework to learn vision-and-language connections is adapted to the glyph-stroke correlation learning, with the expect to increase the structure-awareness of style encoding. In cross-modality pre-training, we build a transformer model that consists of three encoders: a glyph processing module, a stroke processing encoder, and a cross-modality module. Next, to endow our model with the capability of connecting a glyph image and its related stroke labels, we pre-train the model with large amounts of glyph-stroke pairs, via self-supervised signals (reconstruction of the input data). This task helps in learning both intra-modality and cross-modality relationships.

\section{Method}
\subsection{Overall pipeline}

As shown in Figure \ref{fig:framework} (top), the encoder takes two modalities as input: the glyph image of specific style and a sequence of stroke labels representing the corresponding character structure of the glyph. The encoder processes the two modalities separately with two single-modality modules before they are joined with a cross-modality module. In the pre-training stage, the encoder is followed by a convolutional decoder and stroke label predictor, and is designated for self-supervised representation learning (i.e., trained to reconstruct the inputs).

In the second stage, the cross-modality encoder is frozen and used for the downstream task: see Figure \ref{fig:framework} (bottom).  In this stage, we follow the style-content disentanglement paradigm and synthesize novel fonts by combining the style features of the reference glyphs and the content embeddings of the source glyphs. The cross-modality encoder is appended with a decoupling network that aims to decouple the style and content representations from the fused cross-modality representations, which is further followed by a convolutional decoder that is designated to generate novel fonts by taking the style representations of the reference and the content representations of the source as input. 

In the following sections, more detailed descriptions of the sub-modules and training methodologies are presented.

\subsection{Cross-modality encoder}

As shown in Figure \ref{fig:framework}, the cross-modality encoder is made up of two input embedding modules, two single-modality modules and the cross-modality module. The input embedding module converts the input data (glyph and stroke labels) into embeddings sequences. Then, the single-modality module processes each modality separately before they are joined with the cross-modality module. Next, we describe the sub-modules of this cross-modality encoder in detail.

\paragraph{Input embedding module}

The input embedding module converts the input data (i.e., a glyph image and a stroke label sequence) into two separate embedding sequences (glyph embeddings and stroke embeddings). The stroke embedding sequence consists of 28 stroke embeddings that arranged in the stroke order. Note that 28 is the maximum number of strokes forming a commonly-used Chinese character, and the stroke order of each Chinese character is unambiguously determined as strokes are typically arranged based on their spatial coordinates, i.e., from left to right and from top to bottom. The stroke order inputs contain the spatial information, which is also beneficial to improve the final result, especially for the deformation of structure and correct generation of ``connected-stroke''. Each stroke embedding is a sum of three different embeddings: the stroke label embedding, position embedding and modality type embedding. In detail, the label embedding is a 512-dimensional (512-d) vector mapped from the stroke label with an embedding sub-layer. Similarly, the position index (from 0 to 29) of a stroke is projected to a 512-d position embedding with a position embedding sub-layer, and the modality label (0 for stroke and 1 for glyph image) is projected to a 512-d modality type embedding with a modality type embedding sub-layer. Thus, we obtain a 30-embedding sequence for the stroke modality.

On the other hand, the input glyph image of size $256\times256\times3$ is mapped to a feature map of size $8\times8\times512$ with a 5-layer convolutional encoder. The feature map is further flattened to a sequence of 64 512-d embeddings, in which each embedding corresponds to a specific spatial coordinate. Similarly, the position embedding and the modality type embedding of each spatial coordinate are mapped from the x-y coordinate and the modality label respectively with separate embedding sub-layers. The glyph embedding is a sum of the position embedding, modality type embedding and feature embedding. Thus, we obtain a 64-embedding sequence for the glyph modality.

Note that the inclusion of positional information is necessary for the pre-training and font generation task, because the following transformer layers are agnostic to the absolute indices of their inputs as the order of the stroke or image embeddings is not specified.

\paragraph{Self- and cross-attention layers}

We build our single- and cross-modality processing modules mostly on the basis of self-attention layers and cross-attention layers \cite{vaswani2017attention}. After the input embedding module, we obtain two embedding sequences each representing a specific modality. We first apply two single-modality modules, i.e., 9 BERT \cite{devlin2018bert} layers to process stroke information and 5 BERT layers for glyph processing. 

Each cross-modality layer in the cross-modality module consists of two self-attention sub-layers \cite{vaswani2017attention}, one bi-directional cross-attention sub-layer \cite{tan2019lxmert}, and two feed-forward sub-layers. We stack these cross-modality layers in our implementation. The bi-directional cross-attention sub-layer contains two unidirectional cross-attention sub-layers: one from stroke to glyph and one from glyph to stroke. Note that the query and context vectors are the outputs of the former layer (i.e., stroke features and glyph features). The cross-attention sub-layer is used to exchange the information and align the entities between the two modalities in order to learn joint cross-modality representations. For further building internal connections, the self-attention sub-layers are then applied to the output of the cross-attention sub-layer. Lastly, the final output is produced by feed-forward sub-layers. We also add a residual connection and layer normalization after each sub-layer.

\subsection{Pre-training strategy}

In order to learn a better initialization which understands connections between the glyph image and its related stroke labels, we pre-train the cross-modality encoder with a pre-training task on a large font library dataset. As shown in Figure \ref{fig:framework} (top), to ensure effective interaction between the two modalities, during the training phase, there is a probability of 0.375 that all input stroke labels are masked. In the remaining cases, each stroke has a probability of 0.5 to be masked. We attach the encoder with a stroke prediction head consisting of two fully-connected layers. The embedding sequence of the stroke modality is directly mapped to the stroke labels. In addition to where masked strokes are predicted from the non-masked strokes in the stroke label modality, our model could predict masked strokes from the glyph modality as well, so as to resolve ambiguity. For example, as shown in Figure \ref{fig:framework} (top), it is hard to determine the masked stroke from its stroke context but the stroke choice is clear if the visual information is considered. Hence, it helps building connections from the glyph modality to the stroke modality, and we refer to this task as stroke reconstruction task.  We perform the task of learning the labels of masked strokes with cross-entropy loss. We also attach the encoder with a convolutional decoder with the expect to reconstruct the glyph image. The 64-embedding sequence of the glyph modality is reshaped to an $8\times8$ feature map, and then decoded into an image with the decoder. Further, L1 loss contrasting the input glyph and the output image is used to ensure that there is no loss of information.

We aggregate a large aligned glyph-stroke dataset from Founder font libraries \cite{Founder2021}  in the pre-training phase, which consists of 100 different fonts. We pre-train all parameters from scratch (xavier initialization \cite{glorot2010understanding}). Our model is pre-trained with two losses: the glyph reconstruction loss (L1 loss) and the stroke classification loss. We add these losses with equal weights as in Eq. \ref{eqn:loss1}. We take Adam \cite{kingma2014adam} as the optimizer with a linear-decayed learning-rate schedule and a peak learning rate at $1e-4$. We train the model for 30 epochs (i.e., roughly 4,000,000 optimization steps) with a batch size of 4.

\begin{align}\label{eqn:loss1}
L^{pre}=\sum_{i=1}^L BCE(\hat{s_i},s_i) + |\hat{I} - I |
\end{align}
where $I$ is the predicted glyph and $\hat{I}$ is the ground-truth glyph. $s_i, \hat{s_i}$ ($i\in\{1,2,...,L\}$) are the predicted and the ground-truth stroke labels.

\subsection{Downstream task of few-shot font generation}

\paragraph{Model architecture}
Once the encoder is pre-trained, we freeze the parameters and use it for the font generation task. For the downstream task, the encoder is attached to a decoupling network which is made up of 4 Efficient Channel Attention (ECA) modules~\cite{2020ECA} to adaptively rescale channel-wise features and disentangle the style and content representations. The output of the decoupling network is an $8\times8\times512$ feature map, which is split into two $8\times8\times256$ feature maps.  The first split feature map is designated as the style representation and the latter is treated as the content representation. Combining the content representation of the source and style representation of the reference to generate a glyph that represent the source character with the reference style. We employ $8\times8$ feature map instead of a latent vector to preserve richer spatial information. 

\begin{figure}
        \centering
        \includegraphics[width=0.485\textwidth]{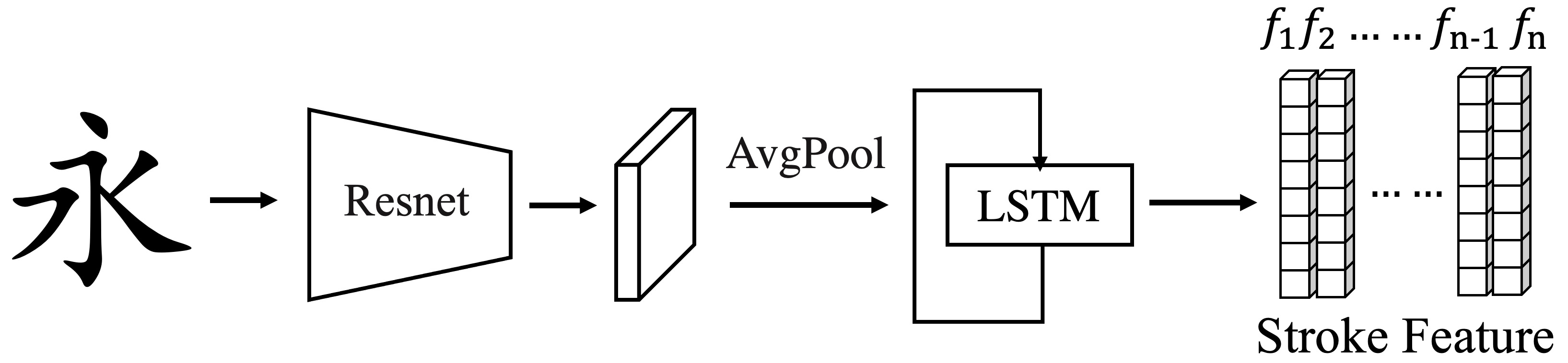}
        \caption[]{The LSTM-based stroke predictor, which is pre-trained for stroke order prediction before used to compute the stroke loss. }
        \label{fig:stroke}
\end{figure}

From the aligned glyph-stroke dataset \cite{Founder2021}, we only use 30 font libraries and 6741 characters from each library in this phase, which are all covered in the pre-training. We train all parameters of the decoupling network and the glyph decoder while holding the encoder parameters frozen. The font generation model is trained with three losses: adversarial loss, reconstruction loss and stroke loss. The adversarial loss encourages generation of valid glyph images by using a discriminator discriminating the generated from ground-truth glyphs \cite{goodfellow2014generative}. The reconstruction loss is the L1 difference between the generated glyph and the exact ground-truth (a glyph of target style and source character). As for the stroke loss, we pre-train an LSTM-based \cite{hochreiter1997long} stroke predictor, which is able to predict the stroke labels sequentially in the correct order given a glyph image as input: see Figure \ref{fig:stroke} for details. Then we use the predictor to compute the stroke loss. Instead of using the predicted labels directly, we employ the activations of the second last LSTM layer and compute the feature differences between the generated and the ground-truth glyph. We add these losses with equal weights as in Eq. \ref{eqn:loss2}. We take Adam \cite{kingma2014adam} as the optimizer with a linear-decayed learning-rate schedule and a peak learning rate at $1e-4$. Like in the pre-training phase, we train the model for 30 epochs (i.e., roughly 5,000,000  optimization steps) with a batch size of 4.

\begin{equation}
    \begin{aligned}\label{eqn:loss2}
    L^{fg} &=Loss_{adv}+ |\hat{I} - I| + |LSTM(\hat{I}) - LSTM(I)|
\end{aligned}\normalsize
\end{equation}
where $I$, $\hat{I}$ are the predicted and the ground-truth glyphs. $LSTM(*)$ is the activations of the second last layer of LSTM-based stroke order predictor, while $Loss_{adv}$ is as the same as WGAN-GP\cite{arjovsky2017wasserstein}.

\section{Experimental Results}

Our model is implemented with PyTorch 1.7 and is trained on a NVIDIA Tesla V100. The pre-training takes 2-3 days and the second-phase training costs 6 days. 

We evaluate the state-of-the-art FFG methods and ours on seen and unseen font domains to measure the generalizability of the models. Our method is compared with five font generation methods on the FFG benchmark, in both the qualitative and quantitative settings. Experimental results demonstrate that XMP-Font achieves the lowest failure rate on both seen and unseen domains. The ablation and analysis study helps understand the role and effects of pre-training strategy, the use of stroke loss and other techniques.

\subsection{Comparisons}

We compared our XMP-Font with two image-to-image translation methods (StarGAN v2 \cite{choi2020stargan} and FUNIT \cite{liu2019few}) and three FFG methods (LF-Font \cite{park2020few}, MX-Font \cite{park2021multiple} and DG-Font \cite{xie2021dg}). StarGAN v2 and FUNIT are not directly proposed for the font generation task, but the universal image-to-image translation paradigm can be applied to the font generation task as well. While \cite{choi2020stargan} only supports translation of glyphs across seen domains, \cite{liu2019few} can be applicable to translations between unseen domains. 

\begin{table}[]
\centering
\setlength\tabcolsep{1.5pt} 
\begin{tabular}{@{}lcccccc@{}}
\toprule
Model   & Few Shot & FID     $\Downarrow$       & PSNR  $\Uparrow$         & SSIM   $\Uparrow$         & L1        $\Downarrow$          & Users   $\Uparrow$        \\ \midrule
\multicolumn{7}{c}{\textbf{Seen} Fonts}                                                                                             \\ \midrule
FUNIT   & Yes      & 147.19         & 8.98           & 0.7069          & 29.38                   & 0.2667          \\
LF-Font & Yes      & 58.90          & 9.78           & 0.7312          & 25.05                   & 0.3725          \\
DG-Font & Yes      & 73.49          & 9.73           & 0.7433          & 25.26                   & 0.5759          \\
MX-Font & Yes      & 66.04          & 9.10           & 0.6963          & 30.05                   & 0.7220          \\
Stargan & No       & 35.24          & 9.82           & 0.7336          & 26.64                   & 0.7974          \\
Ours    & Yes      & \textbf{31.14} & \textbf{12.94} & \textbf{0.7972} & \textbf{19.29}          & \textbf{0.9249} \\ \midrule
\multicolumn{7}{c}{\textbf{Unseen} Fonts}                                                                                           \\ \midrule
FUNIT   & Yes      & 173.30         & 8.45           & 0.6805          & 32.18               & 0.1060          \\
LF-Font & Yes      & 86.33          & 9.35           & 0.7058          & 27.63               & 0.3185          \\
DG-Font & Yes      & 53.04          & 9.33           & 0.7209          & 26.83                  & 0.5220          \\
MX-Font & Yes      & 135.43         & 8.77          & 0.6810          & 26.17                & 0.5906          \\
Ours    & Yes      & \textbf{36.80} & \textbf{12.05} & \textbf{0.7903} & \textbf{18.78} & \textbf{0.8748} \\ \bottomrule
\end{tabular}
\caption{Quantitative evaluations of our XMP-Font and competitors. The reported values are the average of the whole datasets, where only one reference images per style is used for font generation in each experiment.}
\label{tab:comp}
\end{table}

\begin{figure*}
        \centering
        \includegraphics[width=\textwidth]{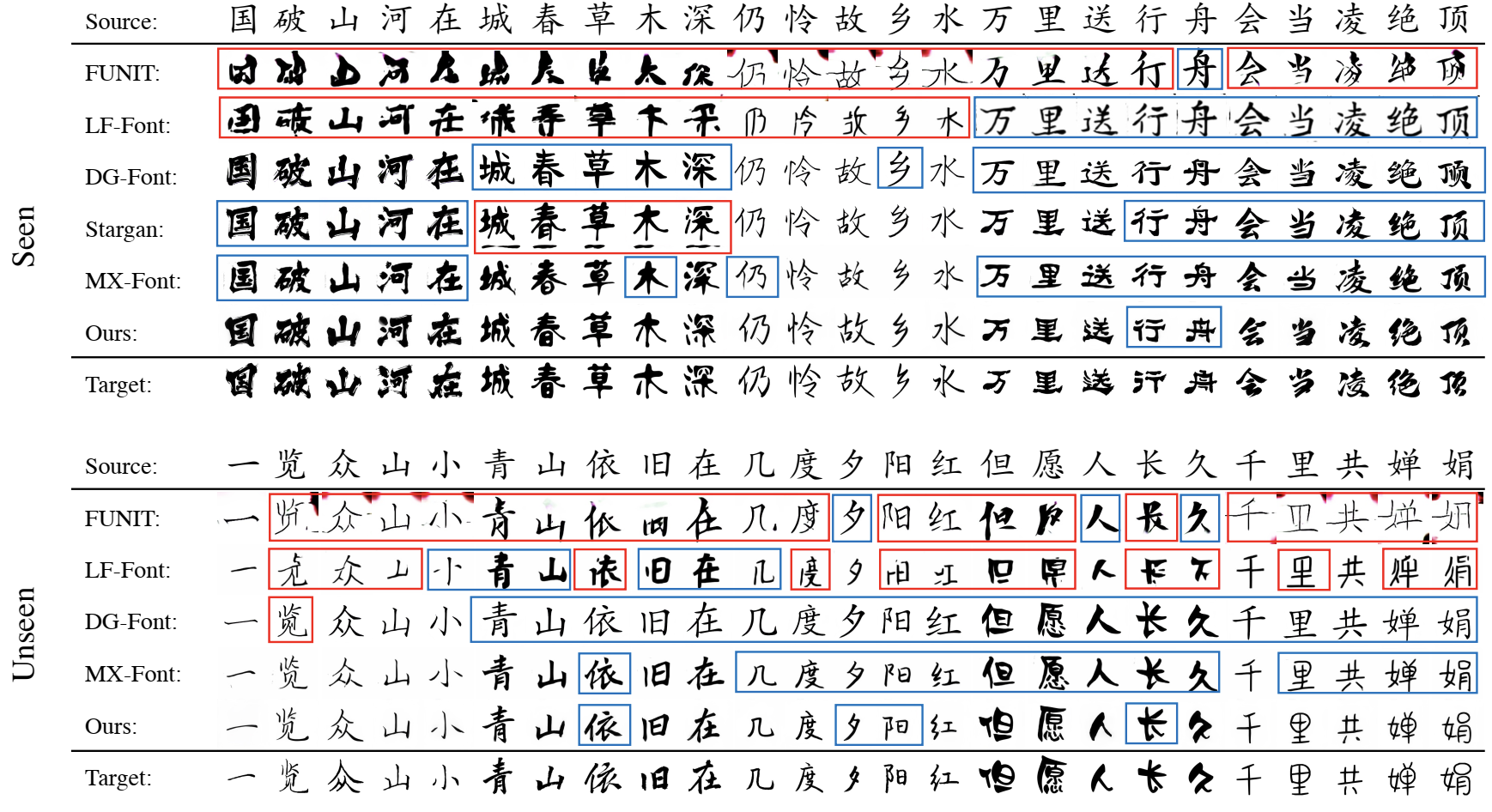}
        \caption[]{Visual comparisons of our XMP-Font with other state-of-the-art methods on famous Chinese poems. The red boxes highlight failures of structure preservation, and blue boxes highlight failures of style transfer.}
        \label{fig:comp}
\end{figure*}

\begin{figure}
        \centering
        \includegraphics[width=0.49\textwidth]{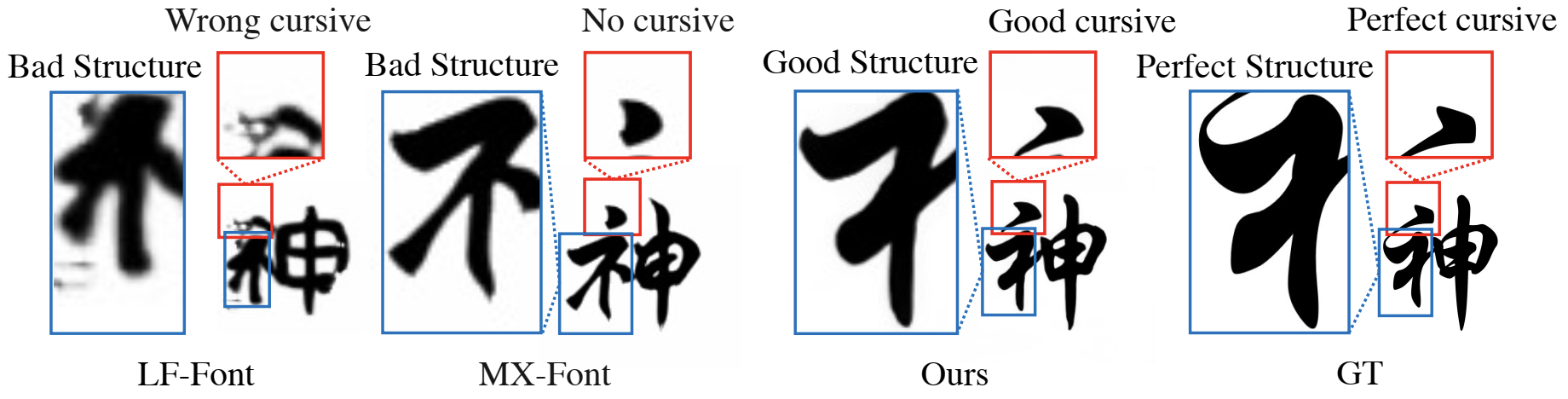}
        \caption[]{Comparisons of our XMP-Font with LF-Font \cite{park2020few} and MX-Font \cite{park2021multiple} in terms of cursive font generation. The glyph details highlighted with the blue boxes and red boxes reveal the noticeable gap between the other two models and ours.}
        \label{fig:comp2}
\end{figure}

To show the generalizability to the unseen style domains, we propose to do the evaluations in the following FFG scenario; training a FFG model on 100 font style domains \cite{Founder2021}, and evaluating the model on both seen and unseen style domains by using only one glyph image as reference. As the stroke labels is independent to font style, stroke labels for all 6741 characters are provided. 

Due to the style of the font domain is defined by appearance features of multiple scales, measuring the visual quality with a unified metric is a challenging problem. As mentioned in MX-Font \cite{park2021multiple}, the multiplicity of the font styles raises the issue when multiple ``ground-truths’’ are satisfying and only one “ground-truth” glyph is present in the evaluation dataset. Thus, in addition to ground-truth-based metrics (SSIM \cite{wang2004image}, PSNR and L1), we also use evaluation metrics that does not require paired ground truths (FID \cite{heusel2017gans}). 

Other than the objective metrics, we conduct a user study for quantifying the subjective quality. The participants are asked to pick the acceptable cases considering the success of style transfer and correctness of the character structure. Failure of either the content or the style  is considered unsuccessful. We randomly select 10 seen font styles and 10 unseen font styles, and 30 characters of each style are generated with each model. Therefore, 3300 samples are generated ($10\times30\times6=1800$ for seen domains and $10\times30\times5=1500$  samples for unseen domains as StarGAN-v2 does not work for unseen domains). The generated samples of the same style and the same character (by different models though) are put together and shown to a participant at a time. The src glyph and a few glyphs of target styles are also shown to the participant in the meantime to facilitate the rating. After all trials finish, the ratings of all participants are collected and analyzed, as presented in Table \ref{tab:comp}. We observe that XMP-Font outperforms other methods in both seen and unseen font generation scenario for most evaluation metrics. In the unseen-domain scenario, ours exceeds others in all metrics by large margin. Especially, our method achieves a remarkably 28\% higher success rate on unseen font domains over the second best. 

We illustrated the generated samples in Figure \ref{fig:comp}. We show the source images in the top row and the corresponding stylize transfer results in the below. In Figure \ref{fig:comp}, we observe that FUNIT\cite{liu2019few} generates the worst results, as it often fails to preserve the character structure of the source (severe loss of strokes or components) and generates unrecognizable glyphs. LF-Font \cite{park2020few} performs well for some test styles, while its performance is unstable as they are prone to loss of strokes or distortion of components on certain style domains. At a glance, other methods including DG-Font \cite{xie2021dg}, StarGAN-v2 \cite{choi2020stargan}, MX-Font\cite{park2021multiple} and ours seem to preserve the character structure well. However, DG-Font \cite{xie2021dg} fails to perform style transfer especially when the source style significantly differs from the target. StarGAN-v2 \cite{choi2020stargan} can only do transfer of seen styles, while it occasionally generates unpleasant stroke paddings as highlighted with the red boxes in Figure \ref{fig:comp}. MX-Font and ours synthesize better detailed structures both in content and style, while there is more chance that MX-Font fails to generate fine-grained style features. 

As shown in Figure \ref{fig:comp2}, the blue and red boxes highlight the failure of LF-Font \cite{park2020few} and MX-Font \cite{park2021multiple} in terms of the generation of stroke-level styles (e.g., selfness), component- and character-level styles (e.g., connected-stroke and inter-component spacing). The advantage of our XMP-Font is highlighted with more successful transfer of styles of all scales. XMP-Font preserves both the detailed local style and fine-grained global styles and generates the plausible and recognizable images consistently. Such a noticeable gap in visual quality explains the large performance leap of XMP-Font in the user study.

\subsection{Ablation studies}

\paragraph{Pre-training strategies}

\begin{figure}[]
        \centering
        \includegraphics[width=.48\textwidth]{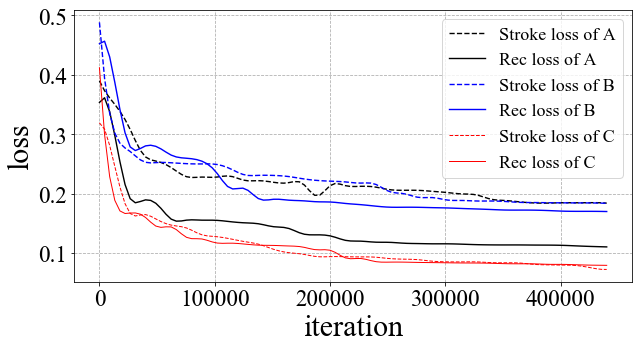}
        \caption[]{The validation losses over optimization steps when using different pre-training strategies. A is the experiment w/o pre-training (the encoder is trained for the downstream task from scratch). B is the experiment with pre-training first and then fine-tuning for the downstream task. C is what is used in our method, namely pre-training the cross-modality encoder first and then keeping it frozen in the second-phase training.}
        \label{fig:ab-loss}
\end{figure}

To show the effectiveness of the pre-training strategy used in our approach, we did three experiments. In Experiment A, the cross-modality encoder is not pre-trained and it is directly trained for the downstream task from scratch. In Experiment B, we pre-train the encoder first and in the second-stage training we only fine-tune the encoder with smaller learning rate ($10^{-6}$). Experiment C is what we use in our approach, namely pre-training the cross-modality encoder first and then keeping it frozen in the second-phase training. Then, we demonstrate the validation losses (stroke loss and reconstruction loss) in Figure \ref{fig:ab-loss}. Figure \ref{fig:ab-loss} clearly shows that Experiment A and B fail to converge as C does, implying that the pre-training strategy is essential for our architecture.

\paragraph{Stroke loss and architecture design}

\begin{figure}[]
        \centering
        \includegraphics[width=.5\textwidth]{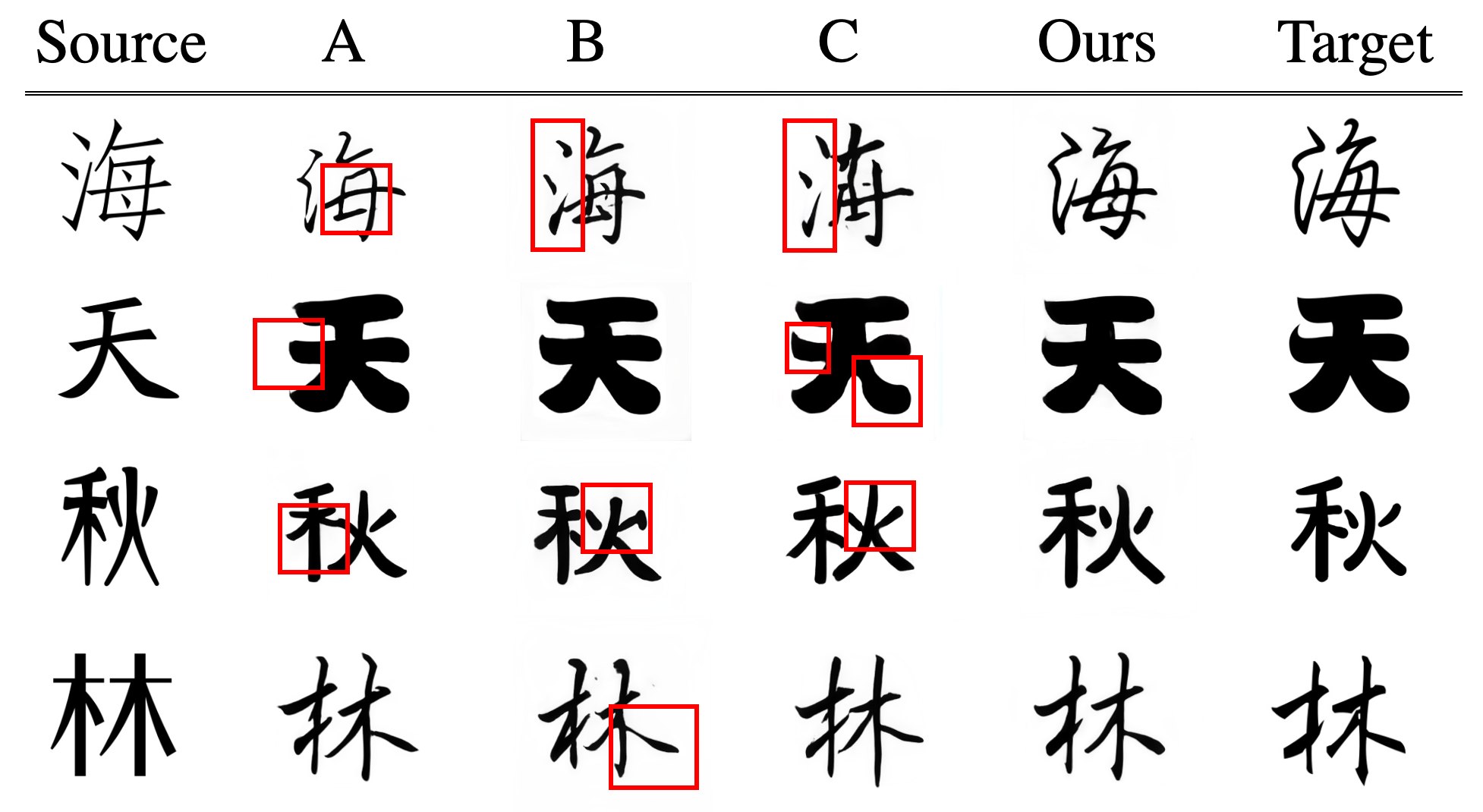}
        \caption[]{Qualitative analysis of the proposed techniques and architecture configurations. A is the experiment w/o stroke loss. B refers to the modified generator architecture with glyph feature size $4\times4$ instead of $8\times8$. C is the modified generator architecture w/o ECA modules \cite{2020ECA}.}
        \label{fig:ab}
\end{figure}

Further, to verify the effectiveness of the proposed stroke loss and the architecture of the generator, we compare the performances of different experimental settings and architecture designs on the validation benchmark under seen and unseen domain transfer scenarios. In Experiment A, we simply remove the stroke loss during the second-stage training. The results are shown in Table \ref{table:ab} and Figure \ref{fig:ab}. Table \ref{table:ab} (A)  shows that the objective metrics worsen without the use of stroke loss. The visual results in Figure \ref{fig:ab} (A) also show that the preservation of character structure becomes worse and the model is less sensitive to the stroke order. 

In Experiment B, we modify the network architecture to adopt smaller glyph feature map ($4\times4$ instead of $8\times8$). To make this possible, we employ a down-sampling convolutional layer upon the $8\times8$ glyph feature map before it is processed with the decoupling network. We observe from Figure \ref{fig:ab}  (B) that the smaller feature map is prone to loss of fine-grained structure information and stroke-level style features. This is also reflected by the objective metrics in Table \ref{table:ab} (B). However, using a $16\times16$ feature map is intractable due to limitation of GPU memory. Therefore, we choose $8\times8$ glyph feature representations in the final architecture.

In Experiment C, we ablate the decoupling network by replacing the ECA modules with a simple convolutional layer. The results in Figure \ref{fig:ab} (C) show that content-style disentanglement worsens without the use of the decoupling network, and some results suffer loss of structure information and incorrect style features.
 
\begin{table}[]
\centering
\begin{tabular}{l|ccccc}
                        & Models                    & A      & B      & C      & Ours   \\ \hline
\multirow{5}{*}{\rotatebox{90}{Seen}}   & \multicolumn{1}{l:}{FID}  & 33.73  & 70.56  & 32.93  & \textbf{31.14}  \\
                        & \multicolumn{1}{l:}{PSNR} & 12.29  & 11.58  & 11.94  & \textbf{12.95}  \\
                        & \multicolumn{1}{l:}{SSIM} & 0.7933 & 0.7689 & 0.7810 & \textbf{0.7972} \\
                        & \multicolumn{1}{l:}{L1}   & 19.85  & 21.25  & 21.07  & \textbf{19.28}  \\
                        & \multicolumn{1}{l:}{Stroke Loss}   & 0.3424 & 0.3896 & 0.3582 & \textbf{0.2709} \\ \hline
\multirow{5}{*}{\rotatebox{90}{Unseen}} & \multicolumn{1}{l:}{FID}  & 41.60  & 56.51  & 45.45  & \textbf{36.80}  \\
                        & \multicolumn{1}{l:}{PSNR} & 11.73  & 10.81  & 11.15  & \textbf{12.05}  \\
                        & \multicolumn{1}{l:}{SSIM} & 0.7725 & 0.7653 & 0.7732 & \textbf{0.7903} \\
                        & \multicolumn{1}{l:}{L1}   & 21.70  & 21.59  & 21.63  & \textbf{18.78}  \\
                        & \multicolumn{1}{l:}{Stroke Loss}   & 0.3692 & 0.4879 & 0.4207 & \textbf{0.3272}
\end{tabular}

\caption{Quantitative analysis of the proposed techniques. A is the experiment w/o stroke loss. B refers to the modified generator architecture with glyph feature size $4\times4$ instead of $8\times8$. C is the modified generator architecture w/o ECA modules \cite{2020ECA}.}
\label{table:ab}
\end{table}

\section{Conclusion}

We proposed the XMP-Font model for few-shot font generation that can generate a novel font library with high success rate by using only one reference glyph from the target domain. Both qualitative and quantitative comparisons with existing methods verify the remarkable advantages of our approach. Our approach significantly boosts the art as it achieves a record-breaking 87.5\% success rate for the few-shot font generation task on unseen font domains. 

Nevertheless, a limitation of our model is that it does not support unseen stroke labels, as it explicitly conditions the style and content representations upon the stroke labels. Neither can it be generalized to unseen languages whose characters are composed of a disparate set of stroke genres.

{\small
\bibliographystyle{ieee_fullname}
\bibliography{zii2i}

\begin{thebibliography}{10}\itemsep=-1pt

\bibitem{azadi2018multi}
Samaneh Azadi, Matthew Fisher, Vladimir~G Kim, Zhaowen Wang, Eli Shechtman, and
  Trevor Darrell.
\newblock Multi-content gan for few-shot font style transfer.
\newblock In {\em Proceedings of the IEEE conference on computer vision and
  pattern recognition}, pages 7564--7573, 2018.

\bibitem{cha2020few}
Junbum Cha, Sanghyuk Chun, Gayoung Lee, Bado Lee, Seonghyeon Kim, and Hwalsuk
  Lee.
\newblock Few-shot compositional font generation with dual memory.
\newblock In {\em Computer Vision--ECCV 2020: 16th European Conference,
  Glasgow, UK, August 23--28, 2020, Proceedings, Part XIX 16}, pages 735--751.
  Springer, 2020.

\bibitem{chen2020uniter}
Yen-Chun Chen, Linjie Li, Licheng Yu, Ahmed El~Kholy, Faisal Ahmed, Zhe Gan, Yu
  Cheng, and Jingjing Liu.
\newblock Uniter: Universal image-text representation learning.
\newblock In {\em European conference on computer vision}, pages 104--120.
  Springer, 2020.

\bibitem{choi2018stargan}
Yunjey Choi, Minje Choi, Munyoung Kim, Jung-Woo Ha, Sunghun Kim, and Jaegul
  Choo.
\newblock Stargan: Unified generative adversarial networks for multi-domain
  image-to-image translation.
\newblock In {\em Proceedings of the IEEE conference on computer vision and
  pattern recognition}, pages 8789--8797, 2018.

\bibitem{choi2020stargan}
Yunjey Choi, Youngjung Uh, Jaejun Yoo, and Jung-Woo Ha.
\newblock Stargan v2: Diverse image synthesis for multiple domains.
\newblock In {\em Proceedings of the IEEE/CVF Conference on Computer Vision and
  Pattern Recognition}, pages 8188--8197, 2020.

\bibitem{Founder2021}
Founder Corp.
\newblock {Founder Type}.
\newblock \url{https://www.foundertype.com/}, 2021.

\bibitem{gao2019artistic}
Yue Gao, Yuan Guo, Zhouhui Lian, Yingmin Tang, and Jianguo Xiao.
\newblock Artistic glyph image synthesis via one-stage few-shot learning.
\newblock {\em ACM Transactions on Graphics (TOG)}, 38(6):1--12, 2019.

\bibitem{gao2020gan}
Yiming Gao and Jiangqin Wu.
\newblock Gan-based unpaired chinese character image translation via skeleton
  transformation and stroke rendering.
\newblock In {\em Proceedings of the AAAI Conference on Artificial
  Intelligence}, volume~34, pages 646--653, 2020.

\bibitem{gatys2016image}
Leon~A Gatys, Alexander~S Ecker, and Matthias Bethge.
\newblock Image style transfer using convolutional neural networks.
\newblock In {\em Proceedings of the IEEE conference on computer vision and
  pattern recognition}, pages 2414--2423, 2016.

\bibitem{heusel2017gans}
Martin Heusel, Hubert Ramsauer, Thomas Unterthiner, Bernhard Nessler, and Sepp
  Hochreiter.
\newblock Gans trained by a two time-scale update rule converge to a local nash
  equilibrium.
\newblock {\em Advances in neural information processing systems}, 30, 2017.

\bibitem{huang2017arbitrary}
Xun Huang and Serge Belongie.
\newblock Arbitrary style transfer in real-time with adaptive instance
  normalization.
\newblock In {\em Proceedings of the IEEE International Conference on Computer
  Vision}, pages 1501--1510, 2017.

\bibitem{huang2020rd}
Yaoxiong Huang, Mengchao He, Lianwen Jin, and Yongpan Wang.
\newblock Rd-gan: few/zero-shot chinese character style transfer via radical
  decomposition and rendering.
\newblock In {\em European Conference on Computer Vision}, pages 156--172.
  Springer, 2020.

\bibitem{isola2017image}
Phillip Isola, Jun-Yan Zhu, Tinghui Zhou, and Alexei~A Efros.
\newblock Image-to-image translation with conditional adversarial networks.
\newblock In {\em Proceedings of the IEEE conference on computer vision and
  pattern recognition}, pages 1125--1134, 2017.

\bibitem{jiang2019scfont}
Yue Jiang, Zhouhui Lian, Yingmin Tang, and Jianguo Xiao.
\newblock Scfont: Structure-guided chinese font generation via deep stacked
  networks.
\newblock In {\em Proceedings of the AAAI conference on artificial
  intelligence}, volume~33, pages 4015--4022, 2019.

\bibitem{johnson2016perceptual}
Justin Johnson, Alexandre Alahi, and Li Fei-Fei.
\newblock Perceptual losses for real-time style transfer and super-resolution.
\newblock In {\em European conference on computer vision}, pages 694--711.
  Springer, 2016.

\bibitem{kingma2014adam}
Diederik~P Kingma and Jimmy Ba.
\newblock Adam: A method for stochastic optimization.
\newblock {\em arXiv preprint arXiv:1412.6980}, 2014.

\bibitem{li2021structurallm}
Chenliang Li, Bin Bi, Ming Yan, Wei Wang, Songfang Huang, Fei Huang, and Luo
  Si.
\newblock Structurallm: Structural pre-training for form understanding.
\newblock {\em arXiv preprint arXiv:2105.11210}, 2021.

\bibitem{li2021few}
Chenhao Li, Yuta Taniguchi, Min Lu, and Shin'ichi Konomi.
\newblock Few-shot font style transfer between different languages.
\newblock In {\em Proceedings of the IEEE/CVF Winter Conference on Applications
  of Computer Vision}, pages 433--442, 2021.

\bibitem{li2020oscar}
Xiujun Li, Xi Yin, Chunyuan Li, Pengchuan Zhang, Xiaowei Hu, Lei Zhang, Lijuan
  Wang, Houdong Hu, Li Dong, Furu Wei, et~al.
\newblock Oscar: Object-semantics aligned pre-training for vision-language
  tasks.
\newblock In {\em European Conference on Computer Vision}, pages 121--137.
  Springer, 2020.

\bibitem{li2017universal}
Yijun Li, Chen Fang, Jimei Yang, Zhaowen Wang, Xin Lu, and Ming-Hsuan Yang.
\newblock Universal style transfer via feature transforms.
\newblock {\em arXiv preprint arXiv:1705.08086}, 2017.

\bibitem{li2018closed}
Yijun Li, Ming-Yu Liu, Xueting Li, Ming-Hsuan Yang, and Jan Kautz.
\newblock A closed-form solution to photorealistic image stylization.
\newblock In {\em Proceedings of the European Conference on Computer Vision
  (ECCV)}, pages 453--468, 2018.

\bibitem{liu2018unified}
Alexander~H Liu, Yen-Cheng Liu, Yu-Ying Yeh, and Yu-Chiang~Frank Wang.
\newblock A unified feature disentangler for multi-domain image translation and
  manipulation.
\newblock {\em arXiv preprint arXiv:1809.01361}, 2018.

\bibitem{liu2019few}
Ming-Yu Liu, Xun Huang, Arun Mallya, Tero Karras, Timo Aila, Jaakko Lehtinen,
  and Jan Kautz.
\newblock Few-shot unsupervised image-to-image translation.
\newblock In {\em Proceedings of the IEEE/CVF International Conference on
  Computer Vision}, pages 10551--10560, 2019.

\bibitem{luan2017deep}
Fujun Luan, Sylvain Paris, Eli Shechtman, and Kavita Bala.
\newblock Deep photo style transfer.
\newblock In {\em Proceedings of the IEEE conference on computer vision and
  pattern recognition}, pages 4990--4998, 2017.

\bibitem{park2020few}
Song Park, Sanghyuk Chun, Junbum Cha, Bado Lee, and Hyunjung Shim.
\newblock Few-shot font generation with localized style representations and
  factorization.
\newblock {\em arXiv preprint arxiv:2009.11042}, 2020.

\bibitem{park2021multiple}
Song Park, Sanghyuk Chun, Junbum Cha, Bado Lee, and Hyunjung Shim.
\newblock Multiple heads are better than one: Few-shot font generation with
  multiple localized experts.
\newblock {\em arXiv preprint arXiv:2104.00887}, 2021.

\bibitem{wang2020eca}
Pengfei Zhu Peihua Li Wangmeng~Zuo Qilong~Wang, Banggu~Wu and Qinghua Hu.
\newblock Eca-net: Efficient channel attention for deep convolutional neural
  networks.
\newblock In {\em The IEEE Conference on Computer Vision and Pattern
  Recognition (CVPR)}, 2020.

\bibitem{DBLPabs210300020}
Alec Radford, Jong~Wook Kim, Chris Hallacy, Aditya Ramesh, Gabriel Goh,
  Sandhini Agarwal, Girish Sastry, Amanda Askell, Pamela Mishkin, Jack Clark,
  Gretchen Krueger, and Ilya Sutskever.
\newblock Learning transferable visual models from natural language
  supervision.
\newblock {\em CoRR}, abs/2103.00020, 2021.

\bibitem{srivatsan2019deep}
Nikita Srivatsan, Jonathan~T Barron, Dan Klein, and Taylor Berg-Kirkpatrick.
\newblock A deep factorization of style and structure in fonts.
\newblock {\em arXiv preprint arXiv:1910.00748}, 2019.

\bibitem{sun2017learning}
Danyang Sun, Tongzheng Ren, Chongxun Li, Hang Su, and Jun Zhu.
\newblock Learning to write stylized chinese characters by reading a handful of
  examples.
\newblock {\em arXiv preprint arXiv:1712.06424}, 2017.

\bibitem{tan2019lxmert}
Hao Tan and Mohit Bansal.
\newblock Lxmert: Learning cross-modality encoder representations from
  transformers.
\newblock {\em arXiv preprint arXiv:1908.07490}, 2019.

\bibitem{tian2017zi2zi}
Yuchen Tian.
\newblock zi2zi: Master chinese calligraphy with conditional adversarial
  networks.
\newblock {\em Internet] https://github. com/kaonashi-tyc/zi2zi}, 2017.

\bibitem{turc2019well}
Iulia Turc, Ming-Wei Chang, Kenton Lee, and Kristina Toutanova.
\newblock Well-read students learn better: On the importance of pre-training
  compact models.
\newblock {\em arXiv preprint arXiv:1908.08962}, 2019.

\bibitem{vaswani2017attention}
Ashish Vaswani, Noam Shazeer, Niki Parmar, Jakob Uszkoreit, Llion Jones,
  Aidan~N Gomez, {\L}ukasz Kaiser, and Illia Polosukhin.
\newblock Attention is all you need.
\newblock In {\em Advances in neural information processing systems}, pages
  5998--6008, 2017.

\bibitem{wang2004image}
Zhou Wang, Alan~C Bovik, Hamid~R Sheikh, and Eero~P Simoncelli.
\newblock Image quality assessment: from error visibility to structural
  similarity.
\newblock {\em IEEE transactions on image processing}, 13(4):600--612, 2004.

\bibitem{wu2020calligan}
Shan-Jean Wu, Chih-Yuan Yang, and Jane Yung-jen Hsu.
\newblock Calligan: Style and structure-aware chinese calligraphy character
  generator.
\newblock {\em arXiv preprint arXiv:2005.12500}, 2020.

\bibitem{xie2021dg}
Yangchen Xie, Xinyuan Chen, Li Sun, and Yue Lu.
\newblock Dg-font: Deformable generative networks for unsupervised font
  generation.
\newblock In {\em Proceedings of the IEEE/CVF Conference on Computer Vision and
  Pattern Recognition}, pages 5130--5140, 2021.

\bibitem{yi2017dualgan}
Zili Yi, Hao Zhang, Ping Tan, and Minglun Gong.
\newblock Dualgan: Unsupervised dual learning for image-to-image translation.
\newblock In {\em Proceedings of the IEEE international conference on computer
  vision}, pages 2849--2857, 2017.

\bibitem{yu2020ernie}
Fei Yu, Jiji Tang, Weichong Yin, Yu Sun, Hao Tian, Hua Wu, and Haifeng Wang.
\newblock Ernie-vil: Knowledge enhanced vision-language representations through
  scene graph.
\newblock {\em arXiv preprint arXiv:2006.16934}, 1:12, 2020.

\bibitem{zhang2021vinvl}
Pengchuan Zhang, Xiujun Li, Xiaowei Hu, Jianwei Yang, Lei Zhang, Lijuan Wang,
  Yejin Choi, and Jianfeng Gao.
\newblock Vinvl: Revisiting visual representations in vision-language models.
\newblock In {\em Proceedings of the IEEE/CVF Conference on Computer Vision and
  Pattern Recognition}, pages 5579--5588, 2021.

\bibitem{zhang2018separating}
Yexun Zhang, Ya Zhang, and Wenbin Cai.
\newblock Separating style and content for generalized style transfer.
\newblock In {\em Proceedings of the IEEE conference on computer vision and
  pattern recognition}, pages 8447--8455, 2018.

\bibitem{zhu2017unpaired}
Jun-Yan Zhu, Taesung Park, Phillip Isola, and Alexei~A Efros.
\newblock Unpaired image-to-image translation using cycle-consistent
  adversarial networks.
\newblock In {\em Proceedings of the IEEE international conference on computer
  vision}, pages 2223--2232, 2017.

\end{thebibliography}


\begin{thebibliography}{10}\itemsep=-1pt

\bibitem{arjovsky2017wasserstein}
Martin Arjovsky, Soumith Chintala, and L{\'e}on Bottou.
\newblock Wasserstein generative adversarial networks.
\newblock In {\em International conference on machine learning}, pages
  214--223. PMLR, 2017.

\bibitem{azadi2018multi}
Samaneh Azadi, Matthew Fisher, Vladimir~G Kim, Zhaowen Wang, Eli Shechtman, and
  Trevor Darrell.
\newblock Multi-content gan for few-shot font style transfer.
\newblock In {\em Proceedings of the IEEE conference on computer vision and
  pattern recognition}, pages 7564--7573, 2018.

\bibitem{cha2020few}
Junbum Cha, Sanghyuk Chun, Gayoung Lee, Bado Lee, Seonghyeon Kim, and Hwalsuk
  Lee.
\newblock Few-shot compositional font generation with dual memory.
\newblock In {\em Computer Vision--ECCV 2020: 16th European Conference,
  Glasgow, UK, August 23--28, 2020, Proceedings, Part XIX 16}, pages 735--751.
  Springer, 2020.

\bibitem{chen2020uniter}
Yen-Chun Chen, Linjie Li, Licheng Yu, Ahmed El~Kholy, Faisal Ahmed, Zhe Gan, Yu
  Cheng, and Jingjing Liu.
\newblock Uniter: Universal image-text representation learning.
\newblock In {\em European conference on computer vision}, pages 104--120.
  Springer, 2020.

\bibitem{choi2018stargan}
Yunjey Choi, Minje Choi, Munyoung Kim, Jung-Woo Ha, Sunghun Kim, and Jaegul
  Choo.
\newblock Stargan: Unified generative adversarial networks for multi-domain
  image-to-image translation.
\newblock In {\em Proceedings of the IEEE conference on computer vision and
  pattern recognition}, pages 8789--8797, 2018.

\bibitem{choi2020stargan}
Yunjey Choi, Youngjung Uh, Jaejun Yoo, and Jung-Woo Ha.
\newblock Stargan v2: Diverse image synthesis for multiple domains.
\newblock In {\em Proceedings of the IEEE/CVF Conference on Computer Vision and
  Pattern Recognition}, pages 8188--8197, 2020.

\bibitem{devlin2018bert}
Jacob Devlin, Ming-Wei Chang, Kenton Lee, and Kristina Toutanova.
\newblock Bert: Pre-training of deep bidirectional transformers for language
  understanding.
\newblock {\em arXiv preprint arXiv:1810.04805}, 2018.

\bibitem{gao2019artistic}
Yue Gao, Yuan Guo, Zhouhui Lian, Yingmin Tang, and Jianguo Xiao.
\newblock Artistic glyph image synthesis via one-stage few-shot learning.
\newblock {\em ACM Transactions on Graphics (TOG)}, 38(6):1--12, 2019.

\bibitem{gao2020gan}
Yiming Gao and Jiangqin Wu.
\newblock Gan-based unpaired chinese character image translation via skeleton
  transformation and stroke rendering.
\newblock In {\em Proceedings of the AAAI Conference on Artificial
  Intelligence}, volume~34, pages 646--653, 2020.

\bibitem{gatys2016image}
Leon~A Gatys, Alexander~S Ecker, and Matthias Bethge.
\newblock Image style transfer using convolutional neural networks.
\newblock In {\em Proceedings of the IEEE conference on computer vision and
  pattern recognition}, pages 2414--2423, 2016.

\bibitem{glorot2010understanding}
Xavier Glorot and Yoshua Bengio.
\newblock Understanding the difficulty of training deep feedforward neural
  networks.
\newblock In {\em Proceedings of the thirteenth international conference on
  artificial intelligence and statistics}, pages 249--256. JMLR Workshop and
  Conference Proceedings, 2010.

\bibitem{goodfellow2014generative}
Ian Goodfellow, Jean Pouget-Abadie, Mehdi Mirza, Bing Xu, David Warde-Farley,
  Sherjil Ozair, Aaron Courville, and Yoshua Bengio.
\newblock Generative adversarial nets.
\newblock {\em Advances in neural information processing systems}, 27, 2014.

\bibitem{heusel2017gans}
Martin Heusel, Hubert Ramsauer, Thomas Unterthiner, Bernhard Nessler, and Sepp
  Hochreiter.
\newblock Gans trained by a two time-scale update rule converge to a local nash
  equilibrium.
\newblock {\em Advances in neural information processing systems}, 30, 2017.

\bibitem{hochreiter1997long}
Sepp Hochreiter and J{\"u}rgen Schmidhuber.
\newblock Long short-term memory.
\newblock {\em Neural computation}, 9(8):1735--1780, 1997.

\bibitem{huang2017arbitrary}
Xun Huang and Serge Belongie.
\newblock Arbitrary style transfer in real-time with adaptive instance
  normalization.
\newblock In {\em Proceedings of the IEEE International Conference on Computer
  Vision}, pages 1501--1510, 2017.

\bibitem{huang2020rd}
Yaoxiong Huang, Mengchao He, Lianwen Jin, and Yongpan Wang.
\newblock Rd-gan: few/zero-shot chinese character style transfer via radical
  decomposition and rendering.
\newblock In {\em European Conference on Computer Vision}, pages 156--172.
  Springer, 2020.

\bibitem{isola2017image}
Phillip Isola, Jun-Yan Zhu, Tinghui Zhou, and Alexei~A Efros.
\newblock Image-to-image translation with conditional adversarial networks.
\newblock In {\em Proceedings of the IEEE conference on computer vision and
  pattern recognition}, pages 1125--1134, 2017.

\bibitem{jiang2019scfont}
Yue Jiang, Zhouhui Lian, Yingmin Tang, and Jianguo Xiao.
\newblock Scfont: Structure-guided chinese font generation via deep stacked
  networks.
\newblock In {\em Proceedings of the AAAI conference on artificial
  intelligence}, volume~33, pages 4015--4022, 2019.

\bibitem{kingma2014adam}
Diederik~P Kingma and Jimmy Ba.
\newblock Adam: A method for stochastic optimization.
\newblock {\em arXiv preprint arXiv:1412.6980}, 2014.

\bibitem{li2021structurallm}
Chenliang Li, Bin Bi, Ming Yan, Wei Wang, Songfang Huang, Fei Huang, and Luo
  Si.
\newblock Structurallm: Structural pre-training for form understanding.
\newblock {\em arXiv preprint arXiv:2105.11210}, 2021.

\bibitem{li2021few}
Chenhao Li, Yuta Taniguchi, Min Lu, and Shin'ichi Konomi.
\newblock Few-shot font style transfer between different languages.
\newblock In {\em Proceedings of the IEEE/CVF Winter Conference on Applications
  of Computer Vision}, pages 433--442, 2021.

\bibitem{li2020oscar}
Xiujun Li, Xi Yin, Chunyuan Li, Pengchuan Zhang, Xiaowei Hu, Lei Zhang, Lijuan
  Wang, Houdong Hu, Li Dong, Furu Wei, et~al.
\newblock Oscar: Object-semantics aligned pre-training for vision-language
  tasks.
\newblock In {\em European Conference on Computer Vision}, pages 121--137.
  Springer, 2020.

\bibitem{li2017universal}
Yijun Li, Chen Fang, Jimei Yang, Zhaowen Wang, Xin Lu, and Ming-Hsuan Yang.
\newblock Universal style transfer via feature transforms.
\newblock {\em arXiv preprint arXiv:1705.08086}, 2017.

\bibitem{li2018closed}
Yijun Li, Ming-Yu Liu, Xueting Li, Ming-Hsuan Yang, and Jan Kautz.
\newblock A closed-form solution to photorealistic image stylization.
\newblock In {\em Proceedings of the European Conference on Computer Vision
  (ECCV)}, pages 453--468, 2018.

\bibitem{liu2018unified}
Alexander~H Liu, Yen-Cheng Liu, Yu-Ying Yeh, and Yu-Chiang~Frank Wang.
\newblock A unified feature disentangler for multi-domain image translation and
  manipulation.
\newblock {\em arXiv preprint arXiv:1809.01361}, 2018.

\bibitem{liu2019few}
Ming-Yu Liu, Xun Huang, Arun Mallya, Tero Karras, Timo Aila, Jaakko Lehtinen,
  and Jan Kautz.
\newblock Few-shot unsupervised image-to-image translation.
\newblock In {\em Proceedings of the IEEE/CVF International Conference on
  Computer Vision}, pages 10551--10560, 2019.

\bibitem{luan2017deep}
Fujun Luan, Sylvain Paris, Eli Shechtman, and Kavita Bala.
\newblock Deep photo style transfer.
\newblock In {\em Proceedings of the IEEE conference on computer vision and
  pattern recognition}, pages 4990--4998, 2017.

\bibitem{park2020few}
Song Park, Sanghyuk Chun, Junbum Cha, Bado Lee, and Hyunjung Shim.
\newblock Few-shot font generation with localized style representations and
  factorization.
\newblock {\em arXiv preprint arxiv:2009.11042}, 2020.

\bibitem{park2021multiple}
Song Park, Sanghyuk Chun, Junbum Cha, Bado Lee, and Hyunjung Shim.
\newblock Multiple heads are better than one: Few-shot font generation with
  multiple localized experts.
\newblock {\em arXiv preprint arXiv:2104.00887}, 2021.

\bibitem{DBLPabs210300020}
Alec Radford, Jong~Wook Kim, Chris Hallacy, Aditya Ramesh, Gabriel Goh,
  Sandhini Agarwal, Girish Sastry, Amanda Askell, Pamela Mishkin, Jack Clark,
  Gretchen Krueger, and Ilya Sutskever.
\newblock Learning transferable visual models from natural language
  supervision.
\newblock {\em CoRR}, abs/2103.00020, 2021.

\bibitem{srivatsan2019deep}
Nikita Srivatsan, Jonathan~T Barron, Dan Klein, and Taylor Berg-Kirkpatrick.
\newblock A deep factorization of style and structure in fonts.
\newblock {\em arXiv preprint arXiv:1910.00748}, 2019.

\bibitem{su2019vl}
Weijie Su, Xizhou Zhu, Yue Cao, Bin Li, Lewei Lu, Furu Wei, and Jifeng Dai.
\newblock Vl-bert: Pre-training of generic visual-linguistic representations.
\newblock {\em arXiv preprint arXiv:1908.08530}, 2019.

\bibitem{sun2017learning}
Danyang Sun, Tongzheng Ren, Chongxun Li, Hang Su, and Jun Zhu.
\newblock Learning to write stylized chinese characters by reading a handful of
  examples.
\newblock {\em arXiv preprint arXiv:1712.06424}, 2017.

\bibitem{tan2019lxmert}
Hao Tan and Mohit Bansal.
\newblock Lxmert: Learning cross-modality encoder representations from
  transformers.
\newblock {\em arXiv preprint arXiv:1908.07490}, 2019.

\bibitem{tian2017zi2zi}
Yuchen Tian.
\newblock zi2zi: Master chinese calligraphy with conditional adversarial
  networks.
\newblock {\em Internet] https://github. com/kaonashi-tyc/zi2zi}, 2017.

\bibitem{vaswani2017attention}
Ashish Vaswani, Noam Shazeer, Niki Parmar, Jakob Uszkoreit, Llion Jones,
  Aidan~N Gomez, {\L}ukasz Kaiser, and Illia Polosukhin.
\newblock Attention is all you need.
\newblock In {\em Advances in neural information processing systems}, pages
  5998--6008, 2017.

\bibitem{2020ECA}
Q. Wang, B. Wu, P. Zhu, P. Li, and Q. Hu.
\newblock Eca-net: Efficient channel attention for deep convolutional neural
  networks.
\newblock In {\em 2020 IEEE/CVF Conference on Computer Vision and Pattern
  Recognition (CVPR)}, 2020.

\bibitem{wang2004image}
Zhou Wang, Alan~C Bovik, Hamid~R Sheikh, and Eero~P Simoncelli.
\newblock Image quality assessment: from error visibility to structural
  similarity.
\newblock {\em IEEE transactions on image processing}, 13(4):600--612, 2004.

\bibitem{wu2020calligan}
Shan-Jean Wu, Chih-Yuan Yang, and Jane Yung-jen Hsu.
\newblock Calligan: Style and structure-aware chinese calligraphy character
  generator.
\newblock {\em arXiv preprint arXiv:2005.12500}, 2020.

\bibitem{xie2021dg}
Yangchen Xie, Xinyuan Chen, Li Sun, and Yue Lu.
\newblock Dg-font: Deformable generative networks for unsupervised font
  generation.
\newblock In {\em Proceedings of the IEEE/CVF Conference on Computer Vision and
  Pattern Recognition}, pages 5130--5140, 2021.

\bibitem{yi2017dualgan}
Zili Yi, Hao Zhang, Ping Tan, and Minglun Gong.
\newblock Dualgan: Unsupervised dual learning for image-to-image translation.
\newblock In {\em Proceedings of the IEEE international conference on computer
  vision}, pages 2849--2857, 2017.

\bibitem{yu2020ernie}
Fei Yu, Jiji Tang, Weichong Yin, Yu Sun, Hao Tian, Hua Wu, and Haifeng Wang.
\newblock Ernie-vil: Knowledge enhanced vision-language representations through
  scene graph.
\newblock {\em arXiv preprint arXiv:2006.16934}, 1:12, 2020.

\bibitem{zhang2021vinvl}
Pengchuan Zhang, Xiujun Li, Xiaowei Hu, Jianwei Yang, Lei Zhang, Lijuan Wang,
  Yejin Choi, and Jianfeng Gao.
\newblock Vinvl: Revisiting visual representations in vision-language models.
\newblock In {\em Proceedings of the IEEE/CVF Conference on Computer Vision and
  Pattern Recognition}, pages 5579--5588, 2021.

\bibitem{zhang2018separating}
Yexun Zhang, Ya Zhang, and Wenbin Cai.
\newblock Separating style and content for generalized style transfer.
\newblock In {\em Proceedings of the IEEE conference on computer vision and
  pattern recognition}, pages 8447--8455, 2018.

\bibitem{zhu2017unpaired}
Jun-Yan Zhu, Taesung Park, Phillip Isola, and Alexei~A Efros.
\newblock Unpaired image-to-image translation using cycle-consistent
  adversarial networks.
\newblock In {\em Proceedings of the IEEE international conference on computer
  vision}, pages 2223--2232, 2017.

\end{thebibliography}
}

\newpage

\end{document}